# Towards Robust Training Datasets for Machine Learning with Ontologies: A Case Study for Emergency Road Vehicle Detection

Lynn Vonderhaar*[1], Timothy Elvira[1], Tyler Procko[1] and Omar Ochoa[1]

*Abstract*—**Countless domains rely on Machine Learning (ML) models, including safety-critical domains, such as autonomous driving, which this paper focuses on. While the black box nature of ML is simply a nuisance in some domains, in safety-critical domains, this makes ML models difficult to trust. To fully utilize ML models in safety-critical domains, it would be beneficial to have a method to improve trust in model robustness and accuracy without human experts checking each decision. This research proposes a method to increase trust in ML models used in safety-critical domains by ensuring the robustness and completeness of the model's training dataset. Because ML models embody what they are trained with, ensuring the completeness of training datasets can help to increase the trust in the training of ML models. To this end, this paper proposes the use of a domain ontology and an image quality characteristic ontology to validate the domain completeness and image quality robustness of a training dataset. This research also presents an experiment as a proof of concept for this method, where ontologies are built for the emergency road vehicle domain.**

*Index Terms*—**Machine learning trust, safety-critical domain, model robustness, model completeness, domain ontology, quality characteristic ontology**

## 1. INTRODUCTION

I N certain domains, Machine Learning (ML) is a way to move towards human-out-of-the-loop decision- making and validating processes. However, in safety-critical domains, humans are still in-the-loop due to a lack of trust in the accuracy and precision of ML [1] [2] [3]. Humans are still heavily involved in data processing and labeling, as well as confirming model decisions because of the catastrophic nature of any incorrect decisions. This issue became even more pronounced in parts of the world after the European Union passed a regulation requiring an explanation to accompany any automated decision using anyone's personal data, causing many companies to revert to human-made decisions [1]. This lack of trust in ML models prevents humans from fully utilizing ML to save time and money, as well as mitigating errors in safety-critical applications, such as autonomous driving [4].

There is plentiful research into methods for explaining the decisions made by ML models. Depending on the nature of the data, explainability methods could include, but are not limited to, various types of heatmaps or linguistic explanations [1]. However, these methods evaluate a specific classification post inference, so each decision must still be reviewed and if the model is part of a pipeline, the inference may have already been passed to a downstream process. While there are also varying metrics to describe model performance, such as accuracy, precision, and recall, these metrics do not capture the significance of incorrect decisions [5]. A more robust approach to building user confidence in an ML model would be applied during design or training so that the trust is for the model, not individual predictions [6]. This work introduces a method for validating the completeness of the training dataset by using dataset quality ontologies to increase user trust in ML models.

There are other areas of ML models that should also be improved to prioritize trust, including design decisions such as hyperparameter tuning. However, ensuring the completeness of the dataset provides the model with the best opportunity to learn unbiased features and patterns. This may help improve user trust in ML models and begin the transition away from human-in-the-loop in safety-critical domains.

The fundamental means of influencing an ML model is through its training set thus the best means of increasing robustness and completeness is also through its training set [7]. If not compiled correctly, a training dataset can introduce bias into the model's decision-making process, among other outcomes such as not converging [6] [7] [8]. Such bias decreases the authority of the model's decisions, whereas an unbiased dataset has a much higher chance of providing unbiased decisions, which could increase user trust [6]. The method presented in this research is a way to reduce bias in training datasets by checking for domain completeness with the help of the Semantic Web. The method also increases model robustness to image quality characteristics by ensuring that examples of varying characteristics are present in the training dataset. The Semantic Web can be used to map ontologies of the desired domain to understand the variation of characteristics in that domain. By comparing each instance of the dataset to a domain ontology, biases existing within the training data can be flagged for improvement. A dataset that only represents a subset of the domain characteristics would be considered biased and would not pass the completeness check; whereas a dataset that represents enough instances of each variation of the mapped domain characteristics would be considered a complete dataset. This research focuses on the identification of emergency vehicles by autonomous vehicles.

[1] Lynn Vonderhaar, Timothy Elvira, Tyler Procko, and Omar Ochoa are with Embry-Riddle Aeronautical University, Daytona Beach, FL 32114 USA (e-mails: vonderhl@my.erau.edu, prockot@my.erau.edu, elvirat@my.erau.edu, ochoao@erau.edu).
* Corresponding author.

## 2. MOTIVATION

There are many suggested solutions for increasing the trust in ML models post training. This section introduces some of them. It is not meant to be a systematic review of all solutions currently available, but merely an introduction to some of the problems with current solutions and why the research presented here is important.

### 2.1. Performance Metrics

There are some metrics already available to assess the confidence of the machine that its decision is correct [5]. Examples include accuracy, precision, and recall, among other metrics [9] [5] [10]. However, each of these metrics has disadvantages that prevent them from providing enough trust to use ML models in safety-critical applications [5] [10]. Accuracy only measures general mistakes but does not note whether the mistakes are false positive or false negative [5]. In other words, if a model predicts the majority class every time, it may have high accuracy, but it is only learning to pick the majority class, rather than learning trends among the classes [10]. In many safety-critical domains, false negatives can result in a worse outcome than false positives, such as an untreated illness or rejecting a good candidate for a loan. Although the accuracy may be near 100 percent, the cost of a mistake that small fraction of the time in a safety-critical application renders the high accuracy meaningless [11]. Therefore, accuracy does not provide necessary background information for the user to trust the decision made. Meanwhile precision and recall do track the false negative rate and false positive rate, but with the opposite problem as accuracy [5] [10]. If a model predicts the minority case every time, it will have high recall but would not be learning the trends between classes [10]. These metrics, along with some others are shown in Fig 1 with the equation for them. In the figure, TP stands for true positive, FP for false positive, TN for true negative and FN for false negative. The figure also provides fairness metrics that correspond with each performance metric. Since F1 score is based on precision and recall, its fairness would correspond with the fairness metrics of those performance metrics.

There are other metrics to describe ML decisions, such as Matthews Correlation Coefficient (MCC), which takes into consideration both false positives and false negatives together but is therefore not considering true positives and true negatives

| Performance Metric | Ideal Value | Fairness Metric | Ideal Value |
|---|---|---|---|
| Recall = TP/P = TP/(TP+FN) | 1 | **Average Odds Difference (AOD)**: Average of difference in False Positive Rates(FPR) and True Positive Rates(TPR) for unprivileged and privileged groups [27]. TPR = TP/(TP + FN), FPR = FP/(FP + TN), $AOD = [(FPR_U - FPR_P) + (TPR_U - TPR_P)] * 0.5$ | 0 |
| False alarm = FP/N = FP/(FP+TN) | 0 | **Equal Opportunity Difference (EOD)**: Difference of True Positive Rates(TPR) for unprivileged and privileged groups [27]. $EOD = TPR_U - TPR_P$ | 0 |
| Accuracy = $\frac{(TP+TN)}{(TP+FP+TN+FN)}$ | | **Statistical Parity Difference (SPD)**: Difference between probability of unprivileged group (protected attribute PA = 0) gets favorable prediction ($\hat{Y} = 1$) & probability of privileged group (protected attribute PA = 1) gets favorable prediction ($\hat{Y} = 1$) [28]. $SPD = P[\hat{Y} = 1|PA = 0] - P[\hat{Y} = 1|PA = 1]$ | 0 |
| Precision = TP/(TP+FP) | 1 | **Disparate Impact (DI)**: Similar to SPD but instead of the difference of probabilities, the ratio is measured [29]. $DI = P[\hat{Y} = 1|PA = 0]/P[\hat{Y} = 1|PA = 1]$ | 1 |
| F1 Score = $\frac{2 * (Precision + Recall)}{(Precision + Recall)}$ | 1 | | |

**Fig 1.** Common performance Metrics. Adapted from *[9]*.

[10]. Additionally, there are probability metrics and metrics based on discriminatory power, such as ROC [10]. Each metric has advantages, but high performance for one metric does not guarantee high performance for another metric, making it difficult to trust a model just based on the mentioned metrics. Therefore, there needs to be a better way to increase user confidence in models.

### 2.2. Explainability Methods

Another method to increase the trustworthiness of ML models is to apply explainability methods to ML decisions. These methods vary with the type of ML model, such as artificial neural networks or natural language processors. These methods can either be black box or white box [12]. In black box methods, the change in the output is analyzed based on the change in the input to understand what part of the input is important to the decision [12]. This leaves the internal workings of the model unknown. Meanwhile in white box methods, the explanation of the decision is developed with an understanding of how the internal model works [12]. These explanations can include words, heatmaps, decision trees, and other ways of describing the decision process [12] [13] [14] [1]. The goal of these explainability methods is to show which parts of the input were important to making the decision [14] [1]. If the proper regions of the input were used to make the decision, it could improve trust that the model can accurately make predictions. However, these methods do not guarantee the safety of the model, but simply describe how a single decision was made. The internal function of the model is still unknown, so proper use of one input does not necessarily prove that the model will properly use and accurately predict all inputs.

### 2.3. The Reject Option

A different method of improving trust in a model is to only allow a model to make a decision that is within its training or above a certain confidence threshold. If it is not sufficiently confident in a decision, it returns no decision [15] [16]. Low confidence could be caused by two rejection types: ambiguity rejection and novelty rejection. Ambiguity rejection is when the model is unable to match the features within the image to a known label [15]. This could be a result of poor image quality or partial occlusion of the object in the image. Novelty rejection occurs when the object in the image is significantly different from the classes that the model was trained on, suggesting that it is a novel class for the model [15].

The reject option can help in improving trust in a model by ensuring that the model will not make a prediction despite low confidence. However, some problems could still arise from using it. The first problem is that epistemic uncertainty shows that a model can be sufficiently confident in an incorrect prediction. If it has not been properly trained for that decision, it may not know its inability to accurately make that decision [17]. Additionally, rejected decisions must be reviewed by the human operator for a decision to be made [15]. This takes additional time before a decision and therefore an action is taken. In some domains, such as autonomous driving, the decision delay, added with the reaction time of the driver could result in an accident, causing injury or loss of life. Therefore, there needs to be a way of ensuring the completeness of a model's training data so that a model is properly trained for each decision within its domain and trained sufficiently to have high confidence in its decisions.

### 2.4. Combinatorial Testing

One way of improving the robustness of an ML model's training data is through combinatorial testing. This has been used to generate test cases for ML models with enough variability of characteristics to represent the domain [18] [19] [20]. Combinatorial testing produces test cases that cover a feasible number of inputs while maximizing the parameter combinations [18] [19] [20]. Given t number of inputs, combinatorial testing is able to generate pseudo-exhaustive test cases, even as t increases significantly, as it does for deep learning models [19] [20]. Although this method can be used to increase trust in ML models by increasing the testing coverage, this does not ensure that the model was built unbiased and trustworthy. Additionally, due to the non-determinism of ML models, testing cannot guarantee that the model is trustworthy.

### 2.5. Subgroup Invariant Perturbation

Another approach to creating trustworthy ML models is subgroup invariant perturbation [8]. This method has been used to reduce bias in decisions by adding a perturbed data term that, when applied to the dataset, transforms its instances [8]. By transforming instances, additional data can be generated for the model to learn on [21]. This method is mathematically different from normal data augmentation but is another way to improve an ML model's robustness to input variations [21]. This perturbed data term, called subgroup invariant perturbation, is based on the estimated bias of the system and model performance [8].

While this method can improve model robustness, there are a couple of problems. First, this is an automated method of data generation, so the black box nature of the ML model is extended even further to include black box data generation [21]. Additionally, the subgroup invariant perturbation term is applied to the provided training dataset, so although additional data is generated, it still does not necessarily cover the entire domain of the data input.

### 2.6. Minimizing Risk and Uncertainty

Some research has been done to quantify and reduce epistemic uncertainty of the model decisions to determine trustworthiness [22] [23] [24]. As this is a result of lack of knowledge, epistemic uncertainty can be reduced by providing the model with more information [23] [24]. An issue with this is that the model does not know classes that it has not been trained on and therefore can still give relatively high confidence for an incorrect prediction, so the uncertainty can sometimes be difficult to detect [17]. Therefore, the dataset that the model trains on needs to be complete to reduce the epistemic uncertainty. This is the method that this paper explores and builds on.

## 3. BACKGROUND

Before describing the experiment performed in this research, some background must be established.

### 3.1. Supervised Learning and Object Detection

Supervised machine learning is a subtype of ML where all instances in the training dataset are labeled. It is often used in classification and regression tasks because the resulting model will provide a specific label as its output [25]. Throughout training, the model will learn patterns and associate it with a hand-curated label, whereas with unsupervised learning, the model has no label for the patterns it learns. In other words, supervised learning attempts to learn the patterns between inputs and their outputs [26].

Object detection is a popular classification task for supervised learning. An object detection model trains on a class or set of classes, then identifies instances of those objects within an input image [27]. There are a few types of deep learning models that are popular for object detection tasks, including Convolutional Neural Networks (CNNs) and Region-based CNNs (R-CNNs) [28] [29]. CNNs are a type of artificial neural network often used in classification tasks due to their ability to filter the most important features in a more computationally efficient way than previous neural networks. The input to a CNN goes through several hidden layers, alternating between convolutional filters and pooling layers, to compute the pixel relevance to pre-defined classes [13]. Although initially used as image classifiers, CNNs can also be used for object detection by using a sliding window across the image and classifying any objects located within that sliding window [28]. However, a faster approach to object detection is with an R-CNN, which generates potential bounding boxes, then uses a classifier within those bounding boxes to refine their locations and dimensions [28] [29]. However, this requires the model to look at the image multiple times to generate and refine the bounding boxes, which still takes a significant amount of time. A more popular model now is the You Only Look Once (YOLO) model, which treats object detection as a regression problem, rather than a classification problem [28]. The model divides the image into a grid and any cell that includes the center of an object must detect that object with a confidence score, as is shown in Fig 2. This is done simultaneously for all objects in the image. In addition to the location, dimensions, and confidence predicted for each grid cell, there is a class-specific confidence score generated for each cell that describes how well the predicted bounding box matches the location and dimensions of the object, which can help refine the bounding box as needed. Throughout this process, the image is only run through the model once, making it much faster than the previous solutions.

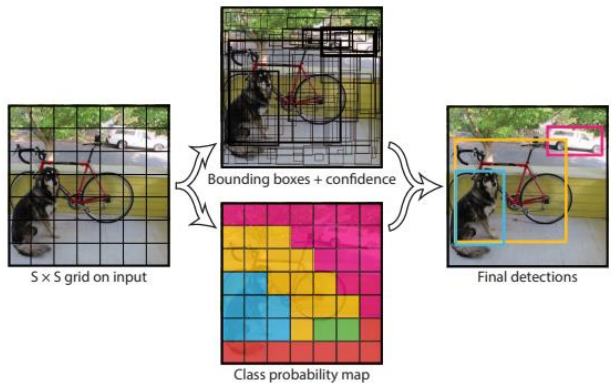

**Fig 2.** YOLO model. Adapted from *[28]*.

### 3.2. Definitions of Safety and Trust

To increase trust in ML models for safety-critical applications, it is necessary to understand both safety and trust. The definition of safety can vary depending on the application, but there are some common underlying factors to every definition. Some define safety simply as the inverse of risk, but others expand the definition to fit a wider scope. For example, safety could be broken down into absolute safety and relative safety [30]. Absolute safety indicates that there is zero risk associated with the model, versus relative safety, which indicates that the risk associated with the model has been reduced to an acceptable level [30]. Complete elimination of risk is not possible in safety-critical applications, so those domains would focus solely on relative safety. Additionally, some definitions of safety separate objective and subjective safety. Objective safety specifies that point of view does not affect the safety of the model, whereas subjective safety specifies that an entity perceives a model to be safe [30]. This definition of subjective safety will be discussed more later as it relates closely to the concept of trust.

Although the definition of safety can vary slightly, there are three main aspects of safety in relation to ML that carry throughout. A safe system will:

1. Minimize risk.
2. Minimize epistemic uncertainty.
3. Prevent severely harmful outcomes resulting from unexpected events [22] [23] [3].

There are several pieces of the definition that need to be broken down. First, to minimize risk. Although there can be some variation by domain, risk tends to be defined using the probability that an event will occur and the severity of harm to the victim when that event occurs [31]. In safety-critical domains, the severity of harm can be serious physical injury, negative impact on the user's livelihood, or even death. Due to the nature of the domains, there is little that can be changed about the severity of a mistake. Therefore, the probability that a model will make a mistake must be reduced as much as possible.

Next, is minimizing epistemic uncertainty. There are two types of uncertainty regarding ML. Aleatoric uncertainty, also known as statistical uncertainty, describes random unexpected events [23] [3]. In other words, it is *irreducible* uncertainty because random events cannot be controlled. Conversely, epistemic uncertainty, also known as systematic uncertainty, can be controlled and is therefore *reducible* uncertainty. It results from a lack of knowledge causing the unexpected event [22] [23] [3]. Since epistemic uncertainty is controllable, reducing it is a key part of producing a safe system. Clearly, uncertainty caused by a lack of knowledge can be reduced by providing more knowledge. In ML, knowledge comes from the training dataset, so high epistemic uncertainty would suggest that the training dataset does not fully encompass the input domain for the model [22]. Further research into the input domain and refining of the training dataset could reduce the epistemic uncertainty of the model.

Finally, a safe system needs to prevent severe harm from befalling the victim when the event occurs. This encompasses the importance of significance of the outcome [22]. It is likely that an unexpected event will occur at some point, no matter how robust the model is. This could be the result of a perturbation too great for the model to overcome, a class outside of the model's training domain, or the model was not exposed to the encountered situation. To ensure a safe ML model, there must be a way to handle these situations that does not result in severe harm to the user. One option for this is a "safe fail" [22] [32]. A safe fail is when the system remains in a safe state, even after failing. Oftentimes for ML, the safe fail is the reject option. If the model's confidence falls below a set threshold, the prediction is rejected, usually by a downstream error detection unit. The user is then prompted to make the decision, such as when a driver must take over control of the vehicle [22] [32].

Together with the completeness of training datasets is the trust that users have in the resulting model. There is a standard view on how to define trust of ML models. The user of the model must be willing to accept a decision or recommendation of a model and act on that acceptance, despite the decision or recommendation having inherent uncertainty [3] [33] [34]. In other words, there is confidence and trust that no harm will befall the user. In terms of ML, users need to trust the model and be willing to be vulnerable to its decisions with the belief that they will not be harmed [3]. The problems raised by this need include how to build a trustworthy model and how to quantify trust so a trustworthy model can be identified after it is built.

### 3.3. Fairness and Bias

Part of the reason that user trust in ML models is low is the bias that models can show when making decisions [6] [35] [8]. Considering the severity of the outcomes in safety-critical domains, the bias decreases trust in the models significantly. In models for job application sorting and criminal justice, it has been shown that biased datasets have led to algorithmic discrimination [36] [37] [38]. This could partially be due to historic data being skewed, such as data for a job application model that would historically indicate men to be the most qualified candidates for engineering positions [36] [39] [37]. After training on a biased dataset, models have been found to amplify the bias found in the training data [38]. To fully accept the use of ML models in safety-critical domains, the bias in training datasets, and by extension, in model decisions, must be reduced.

There are many types of bias in ML datasets; however, this paper focuses on covariate shift, sampling bias, and historical bias [39] [40]. Covariate shift is when a feature of a domain is not sampled completely in the training data [39]. In the example given earlier, this would be the gender feature in the job application model. The training data may have many instances of male engineers, but few instances of female engineers. Sampling bias stems from not randomly taking instances from each subgroup found in the inputs [40]. In this example, this would mean purposefully using male engineers in the training dataset. Finally, historical bias has also been discussed earlier as historically, men have been hired for engineering jobs, so the bias is already present in data [40]. Bias can be classified in many other ways as well, but the common theme is the imbalance of features and labels between the domain and the training dataset. To reduce this bias, all features in the domain would need to be present and correctly balanced in the training data to represent the real-world domain.

Resulting from biased models are unfair decisions. Fairness, defined in decision-making as the lack of bias affecting the decision, can be separated into a few categories: group, individual, and procedural fairness [35] [40]. Group fairness works under the concept that a chosen metric, such as accuracy, is equal across all classes in a set [35] [9]. Individual fairness indicates that similar instances will receive a similar decision [9]. Meanwhile procedural fairness works under the concept that using certain features to decide is inherently fair or unfair [35]. Again, ensuring that each feature and group is represented uniformly in the training dataset could help to reduce model bias and have fair, trustworthy decisions.

### 3.4. Safety Standards

Another reason for the lack of trust in models stems from the software community not having developed standards for ML. For most software systems, there are standards that define safety. For example, part six of ISO 26262 discusses safety of software in road vehicles. It defines the necessary rigor of testing, the amount of documentation needed, and other methods of ensuring safety in such a risky application [41]. An example of ISO 26262 standards regarding error handling is shown in Fig 3. In this example, '++' indicates a highly recommended practice, '+' indicates a recommended practice, and 'o' indicates no recommendation. Additionally, A is the lowest risk application and D is the highest risk application [41]. As seen in the figure, the highest risk systems require the most restrictive standards.

| Methods | | ASIL | | | |
|---|---|---|---|---|---|
| | | A | B | C | D |
| 1a | Static recovery mechanism | + | + | + | + |
| 1b | Graceful degradation | + | + | ++ | ++ |
| 1c | Independent parallel redundancy | o | o | + | ++ |
| 1d | Correcting codes for data | + | + | + | + |

**Fig 3.** Example ISO 26262 standards. Adapted from *[41]*.

While these standards have been successful in maintaining the safety of many systems, many of them were written before the wide acceptance of ML and therefore do not take the difficulties of ML into consideration [42] [43] [44]. For example, testing in safety-critical domains is expected to be highly rigorous according to safety standards, but due to the non-determinism of ML, testing does not indicate the same level of safety as in a normal software system [45]. Additionally, errors in ML decisions could propagate to other aspects of the system, causing the entire safety structure to fail [42]. Therefore, the traditional standards for safety-critical systems need to be amended for systems that incorporate ML, not only in general for software, but also in alignment with standards in other domains. For example, healthcare has safety standards that are also broken by ML, so those standards need to be amended as well before ML models are fully trusted in that domain [46] [47].

There are different organizations trying to incorporate ML into safety standards [48]. For instance, the International Standards Organization (ISO) is working to develop ISO 21448, which covers Safety Of The Intended Function (SOTIF) [42] [49]. This standard recommends that a system should respond in a safe way even if the functionality fails or the user misuses the system [49]. Although this standard and others are being discussed, standards can take years to develop, review, and vote on [50]. As a result, the caliber for safe ML models is still uncertain, making it difficult to use them in safety-critical applications.

### 3.5. Ontologies

An ontology is a structure that defines concepts, or entities, found in a specific domain in reality and relates them to each other [51] [52]. The structure often starts with a taxonomy, or hierarchy, of domain concepts and as more relations are included between concepts, the ontology structure forms. While taxonomies can give valuable information on a domain, the relation between concepts is limited to a subtype relation [51] [53]. In the example of this research, a conventional fire engine *is a* fire truck, which would fit into a taxonomy. However, a conventional fire engine also *has* a ladder, but that relation cannot be modeled in a taxonomy. In contrast, ontologies can model those relations, making their uses more versatile and making it easier for a computer using them to make inferences on information [52]. This difference in capability between taxonomies and ontologies is shown in

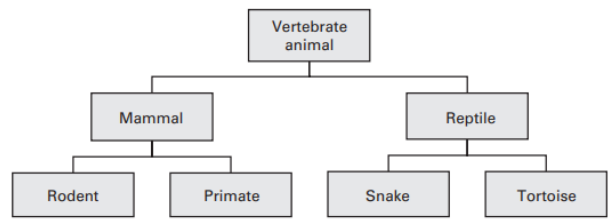

**Fig 4.** Example piece of taxonomy of vertebrates. Adapted from *[51]*.

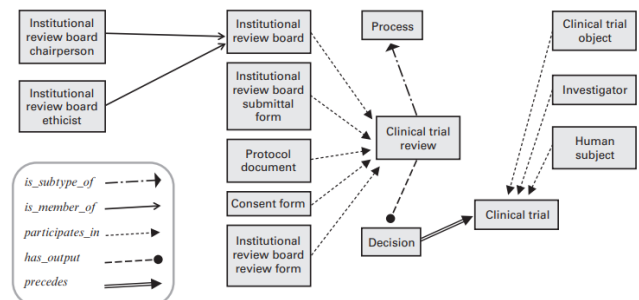

**Fig 5.** Example piece of ontology from an ethical review board. Adapted from *[51]*.

Fig 4 and Fig 5. Fig 4 shows the hierarchical structure of a taxonomy with the single "is a" relation, and Fig 5 shows the versatility in ontology relations and therefore, the versatility in ontology structure.

### 3.5.1. Entities, Universals, Classes, and Particulars

In the context of ontologies, concepts are referred to as entities. An entity can be atomic, or it can be composed of several other entities combined. The combination of entities to form new entities constitutes the definition of a larger entity. For example, the combination of mirrors, a license plate, a windshield, wheels, and a cabin could result in the larger concept of a car. These definitions are what helps to set ontologies apart from taxonomies and what helps machines that use ontologies infer knowledge [52] [53].

Entities are separated into subgroups of entities based on how generalized they are. Universals are the most general entities that represent groups of objects in the real world, but not individual objects. In other words, feature similarities tend to form universals [54]. Universals that are instantiated to represent specific individual objects in the real world are referred to as particulars [52]. Finally, particulars that still represent a group of objects and can be further instantiated, but that do not have a corresponding universal, are known as classes. Additionally, entities are also split based on their number of dimensions. Three-dimensional entities, known as continuants, exist independently of time, such as physical objects. Meanwhile four-dimensional entities, known as occurrents, exist only during a certain period of time, such as events [51] [55] [56].

In ontologies, entities are organized in subject-predicate-object triples where the subject is the starting entity, the object is the ending entity, and the predicate is the relation that points from the subject to the object [57]. These triples help form definitions and attributes of objects, as well as logical constraints that a machine can use to form inferences when using the ontology [58]. Fig 5 shows many examples of the subject-predicate-object triples as every instance of two entities with a connecting edge represents a triple. For example, the section showing decision and clinical trial forms the triple (Decision, precedes, Clinical Trial). This piece of the ontology is isolated in Fig 6.

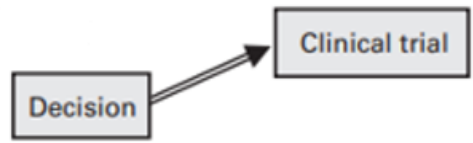

**Fig 6.** (Decision, precedes, Clinical Trial) triple. Adapted from *[51]*.

### 3.5.2. Relations

There are a few core relations in an ontology. One is the relation between two universals, which would represent a relation from the domain taxonomy. This relation symbolizes a parent-child relationship, that one universal *is a* type of another universal. Another core relation in an ontology is one between a universal and a particular, which is a relation of instantiation, which represents going from a general group or type of entity to one specific entity in the real world. This relation allows machines using ontologies to make logical inferences about real world objects [51]. Finally, the relation between two particulars represents composition of one entity from another [51]. These relations are generally found in any high-level ontology structure [51] [59] [56]. However, additional relations can be defined when designing an ontology as a tool to help define the entities.

As stated in the previous section, the relation between two entities forms the predicate of the subject-predicate-object triple. Referring to Fig 5, there are several types of relations between the entities. The list of these entities is shown in a legend, which is repeated in Fig 7. The legend shows that although there are several instances of the *is_subtype_of* (or simply *is_a* as it has been referred to in this section), which forms the backbone of the ontology. However, there are also other relations, such as *has_output* and *participates_in* which are more specific to that ontology.

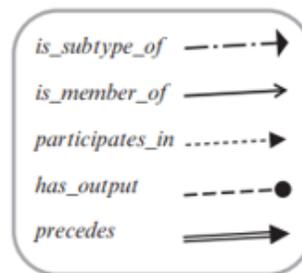

**Fig 7.** Legend of relations. Adapted from *[51]*.

### 3.5.3. High Level Ontologies

One issue that was discovered with ontologies was the lack of standardization in structure as various people implemented their own ontologies [51]. To try to reduce this problem, some high-level ontologies, also known as formal ontologies, were introduced with the goal of providing a more standard structure for others to emulate as they designed an ontology for their own specific domain. High-level ontologies are domain independent, so they can be reused to structure ontologies in multiple domains [60]. They help to maintain the domain taxonomy backbone, as well as separate handling of continuants and occurrents [51]. Additionally, any rules, such as taxonomy transitivity (if a conventional fire engine *is_a* fire truck and a fire truck *is_a* fire vehicle, then a conventional fire engine *is_a* fire vehicle), are enforced when using a high-level ontology [51].

There are a few commonly used top-level ontologies that can be extended to structure domain ontologies. One is the Suggested Upper Merged Ontology (SUMO), which describes entities, relations, predicates, subclasses, instances, and more. SUMO can be used as a guide for designing knowledge bases, as well as domain ontologies. It has 11 sections that cover topics such as units of measure, set theory, and ontology structure [61] [62] [63]. SUMO was built by combining several top-level ontologies to form universals out of the common features of their universals. The terms are therefore quite broad, so SUMO can be used in a large variety of domains.

Another commonly used top-level ontology is the Basic Formal Ontology (BFO), which is generally used in scientific domains and is approved by the ISO [64] [65]. BFO is formatted to classify entities as either continuants or occurrents and to format their attributes based on that classification. Top-level ontologies use even more general universals to avoid any domain knowledge entering their scope [51]. They are meant to be extended by a domain ontology with an entity that *is_a* continuant or an entity that *is_a* occurrent. By making that extension, that entity then gains each of the subtypes of continuant or occurrent that will guide the development of the entire domain ontology. BFO is the top-level ontology that this research focuses on.

### 3.6. Knowledge Graphs

While ontologies represent knowledge in domains in a broad sense, without including individual instances of the entities, knowledge graphs extend ontologies to include those individual instances. Knowledge graphs follow the same structure as ontologies with subject-predicate-object triples. They also are based on an ontology, meaning that the ontology structures the knowledge in the domain and describes relations between entities, and a knowledge graph for that domain would instantiate the entities found in the ontology [66] [67] [68]. In the example given thus far, the ontology may include the entity conventional fire engine, which is a general group of objects in the real world that have a manufacturing year, a color, a license plate number, and other attributes. Meanwhile in a knowledge graph of the domain, that entity could be instantiated to a specific 2009 red fire engine with a defined license plate number. This concept is represented again in Fig 8, which is a knowledge graph describing Albert Einstein [67]. An ontology for this knowledge graph could relate a scientist to their parents, where they were born, their theory, their supervisor, and so on. The knowledge graph then shows a specific instance.

Ontologies and knowledge graphs can both be queried to elicit knowledge, the use of one or the other depends solely on the application and purpose that they are used for. For data usage only requiring classes of data, an ontology is sufficient, whereas for data usage requiring specific individual cases, a knowledge graph is necessary. The research presented here will utilize parts of the ontology as is for the domain knowledge, while knowledge on image quality characteristics will be instantiated.

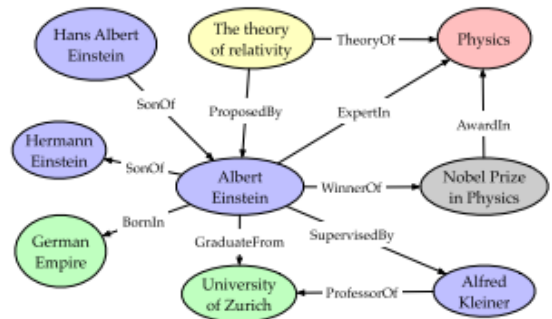

**Fig 8.** Knowledge graph of Albert Einstein. Adapted from *[67].*

### 3.7. Querying

Once the knowledge is organized in an ontology or knowledge graph, there needs to be a way of accessing it and formatting the results. There are several languages for querying, or searching, ontologies and knowledge graphs. These languages are used to query ontologies and knowledge graphs in Resource Description Framework (RDF) format, much like the Structured Query Language (SQL) is used to query relational databases. One such language is the SPARQL Protocol RDF Query Language (SPARQL), which can be used to query data in RDF, XML, JSON, and other formats [69] [70]. Primarily, this means that SPARQL can query data formatted into subject-predicate-object triples. An example SPARQL query can be seen in Fig 9. This query is asking for all emails related to the entity Craig.

```
SELECT ?craigEmail
WHERE
{ ab:craig ab:email ?craigEmail . }
```

**Fig 9.** Example SPARQL query. Adapted from *[69].*

Another way of looking at a SPARQL query like the one seen in Fig 9 is shown in Fig 10. This depiction shows that the WHERE command selects the entity or group of entities that the query is looking at and the SELECT command then takes the specified attributes of those entities. When considering the example from Fig 9, the WHERE command selected Craig and the SELECT command selected email.

SPARQL is not the only query language for ontologies. There are other query languages that are used for ontologies in non-RDF format. However, SPARQL is a standard query language for RDF ontologies, so that is the language choice that this research will use [71].

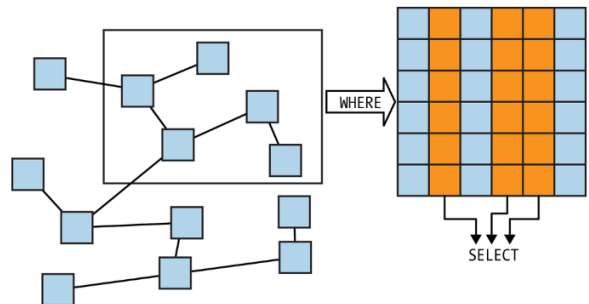

**Fig 10.** WHERE and SELECT in SPARQL queries. Adapted from *[69]*.

### 3.8. The Semantic Web – Goals, Formatting, and Current Access Portals

The semantic web is the machine-readable version of the world wide web. Information on the world wide web is encoded in natural language, tables, images, videos, and many other formats that make the knowledge easier for humans to understand but make it impossible for machines to dissect the information found there [72]. The goal of the semantic web is to structure knowledge found on the world wide web into ontologies and knowledge graphs that can be queried by machines to infer more meaningful knowledge based on entities and relations. In other words, it allows machines searching the web to understand the meaning, or semantics, of what they find there [69] [72]. The semantic web is based on RDF, so the information found there is organized in subject-predicate-object triples and can be queried using SPARQL.

The semantic web was first introduced by Tim Berners-Lee and was pitched as an extension of the world wide web that would streamline knowledge access and create more meaningful knowledge. Therefore, the goal was to have all information in the world wide web also represented in the semantic web [72]. Unfortunately, the semantic web fell short of this goal, but there are a few portals that allow humans to access web data organized in RDF. For example, KBpedia and DBpedia are both compilations of large ontologies that can be queried to gain meaningful information from the web [73] [74]. KBpedia was chosen for the research presented here because of its ease of use and its larger knowledge base. This semantic web access portal has combined knowledge from the following seven knowledge bases:

1. Wikipedia for a general knowledge base.
2. Wikidata for several million entities represented in knowledge graphs.
3. Schema.org for additional knowledge graphs on common sense and logic.
4. DBpedia for some additional Wikipedia knowledge represented in RDF.
5. GeoNames for geographical data.
6. OpenCyc for more information on common sense and logic.
7. UNSPSC products and services for information regarding eCommerce.

KBpedia is freely available on the Internet and can be queried to help define domains and build domain ontologies [74]. Its RDF structure makes the transfer of entities and relations from the knowledge base to an ontology simple.

### 4. ONTOLOGY DEVELOPMENT

Before describing the experiment, it is important to establish the significance of dataset completeness and the development of the ontologies.

### 4.1. Significance

As discussed previously, the completeness of the training dataset can be a step towards building user trust in the model [7]. Since most of the model's complexity is built from the training data, biases can be embedded into the model via biased data [7] [6] [8]. In some applications, this bias can be a mere nuisance. However, in safety-critical applications, biased decisions can have life or livelihood threatening consequences, so model bias in those applications needs to be minimized [7]. There are many types of bias that can be introduced to the model, but this research focuses on one introduced by unrepresentative domain data [75]. Therefore, for this purpose it is important to ensure the domain completeness of a training dataset, as well as the correct percentage of representation of each class in the domain to avoid representation bias.

In addition to domain bias, it is important in safety-critical applications to ensure the robustness of the model to quality characteristics in images [76]. Models trained only on clear images have a significant performance decrease when facing lower quality images [77] [78]. If an autonomous vehicle has not been trained on any images of vehicles blurred by movement, then seeing motion blur in the real application could cause it to misclassify an object. In many cases, models are not only trained solely on clear images, but they are also validated on only clear images [79]. This introduces a false sense of impressive model performance that can be shattered when the model is used on imperfect data. It is essential, therefore, that lower quality images be included in the training dataset.

Clearly, both domain completeness and quality characteristic robustness must be addressed within the training dataset for ML models to improve user confidence in models used for safety-critical applications. Domain completeness is necessary to reduce representation bias and robustness is necessary to improve model performance on imperfect data.

### 4.2. Domain Completeness

Mentioned previously, there are numerous causes of bias in ML models, one major cause is an unrepresentative training dataset, which is the cause of bias that this research focuses on. Unrepresentative means that training instances are either entirely ignoring some classes in the relevant population or underrepresenting the percentage of the relevant population that a class represents [80] [81]. A training dataset tends to cover a subset of the total data population. This is shown in Fig 11, where the bottom oval represents the entire population that relates to the application, the cylinder shows the subset of the relevant population that is used to train the model, and the model that is trained then affects the relevant population through its decisions [75]. If not carefully chosen, that subset of data could be unrepresentative of the total population, causing the training and resulting decisions of the model to be biased [80] [81]. The class distribution of a training dataset will affect the training of the ML model. Therefore, a class that does not appear in the training data will almost certainly not be chosen outside of training. Additionally, underrepresenting a certain class in training will make the model less likely to choose that class outside of training, which would bias the model [80] [82]. Finally, as seen in Fig 11, biased decisions could affect who or what is in the relevant population, thereby making the bias even more pronounced [75]. It is therefore critical that the training data be representative of the entire relevant population to avoid representation bias in the model as unrepresentative data could set off a cycle creating an increasingly biased model.

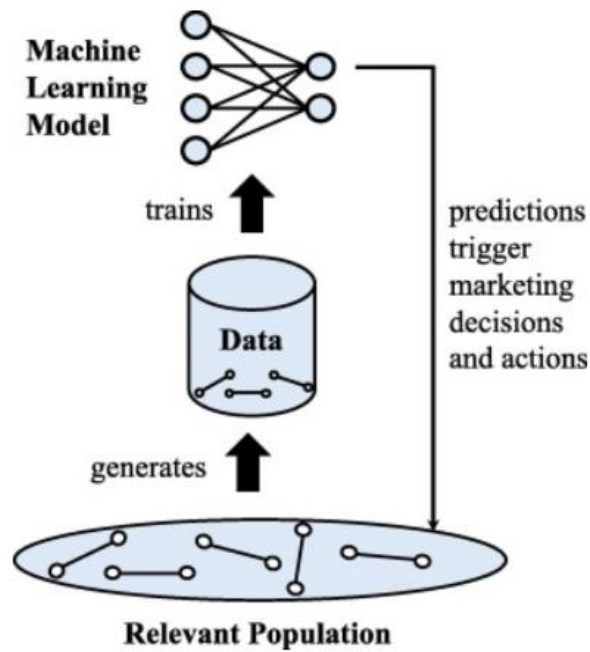

**Fig 11.** Generating training data. Adapted from *[75]*.

It can be difficult to ensure the complete coverage of a domain as sometimes domain boundaries are subjective. Defining the scope of a domain is a common difficulty in ontology engineering, but it is a critical first step to designing the ontology [83] [84]. Missing entities or including incorrect entities could introduce bias as well because the relevant population, and potentially the representative population for training the model, does not truly represent the domain. However, the research presented in this paper utilizes the semantic web to help define the boundaries of the domain and ensure that the domain coverage is complete. The semantic web has the advantage of massive knowledge bases that are already structured as ontologies. This makes building a comprehensive domain ontology much less time consuming and helps to avoid boundary mistakes.

Another difficulty with minimizing representation bias is managing the dataset percentage representation. Since misrepresenting proportions in datasets can lead to bias, it is important to maintain proper class proportion within a training dataset [81]. Since probability plays such a large role in ML decisions, proportion misrepresentation can greatly decrease model performance and therefore, user trust in models, especially when consequences involve lives and livelihood. Therefore, domain completeness, as defined in this research, includes both the complete coverage of classes within a domain, as well as proper representation of proportions of those classes within the domain's relevant population.

### 4.3. Quality Characteristic Robustness

In addition to domain completeness, it is important in safety-critical applications that the trained ML model be robust to image quality degradation, also known as perturbation or distortion. When in use, the model cannot be sure that an image fed to it will be perfect. In safety-critical applications, the user must be able to trust that a slight blur or slightly low contrast will not void the model's decision. There are multiple methods to increase model robustness including a denoising autoencoder which removes the perturbations and low-quality characteristics from an input, as well as dropout, which drops pixels randomly during training to simulate perturbations [85] [86]. The method used in this research adds necessary image quality characteristics to the training dataset to ensure robustness. This method includes examples of lower quality images in the training dataset so that the model can handle those characteristics when faced with data after training. The method is similar to one known as adversarial training, which includes common adversarial attacks against models in the training dataset [85] [87]. However, rather than including small perturbations that would indicate an adversarial attack, this research focuses on simulating low image quality resulting from normal operational mistakes. However, it can easily be expanded to address other concerns, such as adversarial attacks.

There are many types of perturbations that are possible in images. One study revised the ImageNet dataset to reflect fifteen types of image perturbations, as shown in Fig 12. This study also separates the perturbations by five severity levels [76]. Another study tested model performance after fine tuning a model with varying types of perturbations [88]. The perturbations chosen for this research will be discussed further in later sections. Those necessary for safe training may vary between domains depending on the application and the environment of the system. For example, images including fog may be necessary for an autonomous vehicle model, but not a medical diagnosis model. However, the concept is the same as domain completeness. The training data must reflect each of the quality degradation types with a correct percentage to accurately represent the relevant population and minimize model bias. The semantic web knowledge base used in this research did not cover perturbation types, so separate research had to be conducted to determine the boundary of that domain for the developed ontology.

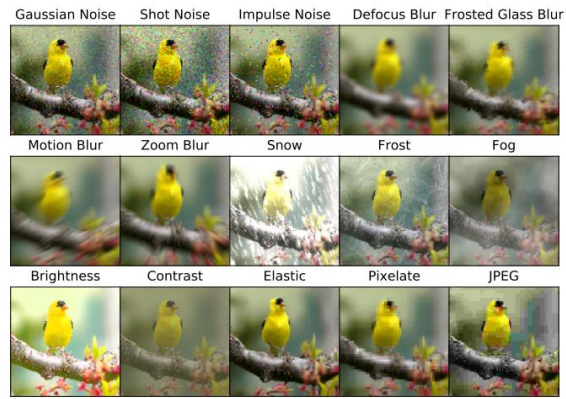

**Fig 12.** Types of perturbations chosen by one study. Adapted from *[76]*.

### 4.4. Ontology Development

With the important aspects of the training data considered, the ontologies for this research could be built. There are three ontologies to cover domain knowledge, image quality knowledge and ML knowledge. The domain knowledge covers emergency road vehicles. However, the structure of the developed ontologies makes expanding the domain quite simple.

The ontologies were built using Stanford University's Protégé ontology builder [89]. Protégé is the current standard for ontology development. It allows developers to add entities with a hierarchical structure starting from the standard owl:Thing entity. The hierarchical structure enforces the is_a relation among the entities in the ontology. From there, developers can add varying relations between entities, which turn the structure from a taxonomy to an ontology. All of the entities and relations can then be visualized as a graph.

For this research, the entities from the semantic web were related using relations from BFO. More specifically, the domain ontology uses the "contained in" and "part of" relations, as well as their inverses "contains" and "has part". Since BFO is an upper-level ontology, using its relations makes the expansion and reuse of the domain ontology much easier [90]. Upper-level ontologies, such as BFO, are well-known and widely used, so aligning this domain ontology with BFO enables any other ontology also aligned with BFO to be able to use this domain ontology.

It is important to note that each class within the training data should have instances of each perturbation type. Research has compared the performance of models trained on training sets of varying degrees of clarity and perturbation. The results, shown in Fig 13, indicate that the model will fit to its training data, so if trained on only clear images, it will only perform well when classifying clear images. Similarly, if only trained on noisy or blurry images, it will perform well on clear images and images with that type of perturbation, but not other types of perturbations. Finally, when trained on a mixture of clear images and images with varying types of perturbations, the model performed well when classifying all three types of images (clear, noisy, and blurry) [88]. Therefore, the research presented in this paper suggests including various types of distortions and at various levels to each domain class to maximize the model's performance under many circumstances. Ideally, improving model performance when facing various classification obstacles will begin to increase user confidence in ML models for safety-critical applications.

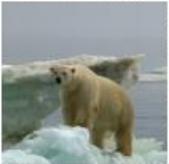
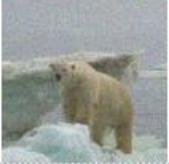
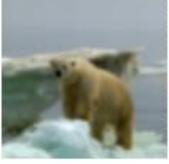

**Fig 13.** Performance of models trained on varying levels of clarity. Adapted from *[88]*.

### 4.5. Domain Ontology

Ensuring the training dataset has full domain coverage is critical to the success of this approach; however, it is challenging to ensure full domain coverage. Therefore, the base of the domain ontology was formed using the semantic web to help define the boundaries of the domain. There are many possible knowledge bases to choose from for the semantic web. The knowledge base chosen for this research was KBpedia because it combines knowledge from several other bases.

The KBpedia knowledge base provided the backbone taxonomy of safety and rescue vehicles for the domain ontology. The entities represented in KBpedia did not fill the entire ontology because while they included broad categories of vehicles, such as "ambulance", they did not provide more detailed sub entities, such as "rapid organ recovery ambulance" or "non-transporting EMS vehicle". The knowledge bases that built KBpedia were explored further to develop and expand the domain ontology. KBpedia has links to the Wikipedia knowledge base, which provided information on additional types of vehicles for the rest of the domain ontology. The references for each of these domain entities can be found in the Appendix. In addition to the taxonomy of rescue vehicles, some other entities were included to help describe the emergency vehicles, such as Emergency_Light. In the domain ontology, the general category of safety and rescue vehicles is split into the following six more specific categories: Emergency Medical Services (EMS) vehicles, fire vehicles, mobile communications vehicles, police vehicles, rescue vehicles, and tow trucks. This can be seen in Fig 14. Also shown in Fig 14, is how a Safety_Rescue_Vehicle relates to upper-level aspects of an image, which allows the domain ontology to align with the other two ontologies in this process. A Safety_Rescue_Vehicle is a Subject, which is part of a Label. An Image consists of both a Subject and a Label.

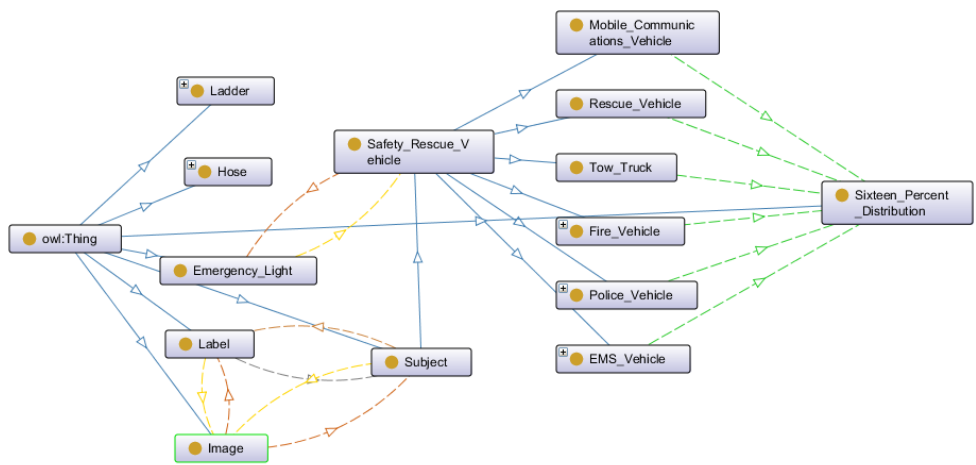

**Fig 14.** The split of Safety_Rescue_Vehicle.

The six categories of Safety_Rescue_Vehicle are then expanded, when applicable, to even more specific types of each of the vehicles. The expansions of the domain ontology can be seen in the Appendix. As shown in the ontology, each of the six categories of rescue vehicles has an intended distribution within the dataset that must be matched to pass this test.

### 4.6. Image Quality Characteristics Ontology

The second part of the process is to ensure the training dataset has complete coverage of image quality characteristics. The contents and boundaries of this ontology proved to be more difficult to define than those of the domain ontology because KBpedia did not contain entities related to image quality. Additionally, upon further research, no other image quality ontologies were found as a knowledge base. Therefore, the knowledge base for this ontology was academic papers found using Google Scholar.

With emergency vehicles being the domain of this experiment, the image quality characteristics and values chosen center around quality characteristics that can occur outdoors. The quality characteristics in the ontology are blur, contrast, illumination, occlusion, and resolution. Blur is then expanded into four types: defocus blur, gaussian blur, haze blur, and motion blur. Each of these types of blurs were also chosen with emergency vehicles and autonomous driving in mind. Defocus blur occurs from a camera not focusing properly, gaussian blur is caused by turbulence in the atmosphere, haze blur can occur from fog, and motion blur can be caused by the relative motion of the vehicles. In relation to the experiment, a training dataset must have varying levels of each of those characteristics to be considered complete.

Fig **15** shows the expansion of the quality characteristics. As with the domain ontology, the quality characteristics are also related to upper-level aspects of an image. The characteristics are contained in Labels and are part of an Image.

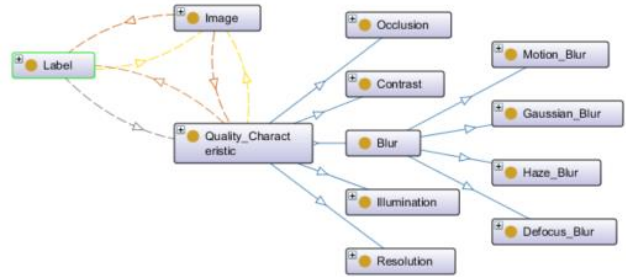

**Fig 15.** Image Quality Characteristics.

### 4.6.1. Value Selection and Ontology Individuals

One major difference between the domain ontology and the image quality characteristic ontology is that each entity representing a quality characteristic had to be expanded to define different ranges that a training dataset needed to include. In Protégé, these ranges can be represented as individuals that partially instantiate the entities. Fig 16 shows the expansion of Occlusion into four individuals: Occlusion_None, Occlusion_Low, Occlusion_Medium, and Occlusion_High. Each of the other quality characteristics were expanded in a similar fashion. The individuals were then defined as ranges of values to provide exact information on what the training dataset must include. The full list of individuals, their value range, and the references for their values can be found in the Appendix. Each of the individuals has a property that includes their intended distribution.

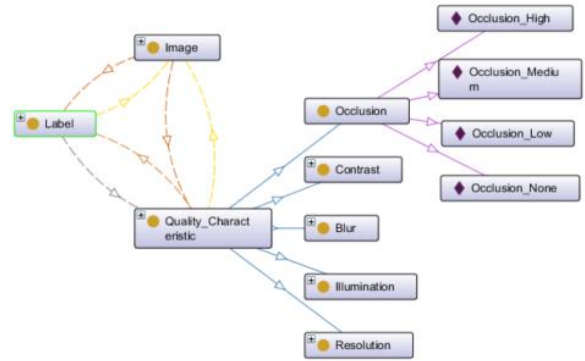

**Fig 16.** Occlusion Individuals.

### 4.7. Machine Learning Model Ontology

The final ontology for this experiment is one that describes ML models. The knowledge base for this ontology is the ML Schema ontology [91]. However, the ontology was trimmed in some places and expanded in others from the ML Schema ontology to align with the experiment and the other two ontologies. The basic structure of this ontology is shown in Fig 17. Each of the entities expand to sub entities and relations to other entities, but the most important relations to discuss here are those related to Data and ModelEvaluation, which are shown in the Appendix.

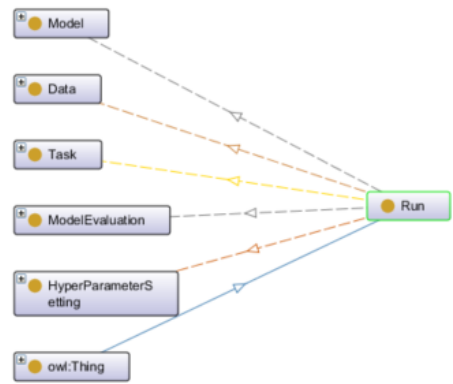

**Fig 17.** Basic taxonomy of an ML model.

## 5. EXPERIMENT

The domain used in this experiment is object detection of emergency road vehicles. Since the research focuses on the completeness of ML training datasets in safety-critical applications, the experiment uses a safety-critical domain. This domain simulates an application towards autonomous driving, when emergency vehicles would need to be recognized for resulting behavior, such as moving over and stopping.

### 5.1. Dataset Compilation

In total, there were three datasets to complete the experiment: a full dataset that passed through the process, a dataset that was missing a domain entity that was flagged via SPARQL queries, and a dataset that was missing an image quality characteristic that was flagged via SPARQL queries.

All of the images used in these datasets were found using a Google Images search for "emergency vehicle", or for individual entities from the domain ontology. The full list of references for these images can be found in the Appendix. To maintain a high level of control over the dataset and image quality values, many of the image quality characteristics for this experiment were added manually to clear, high-quality images. While this was unnecessary for some characteristics, such as nighttime illumination, as that was simple for a human to recognize and annotate, more difficult characteristics, such as the varying types of blurs, were added to clear images. By controlling the transformations applied to the images for blur, contrast, and illumination, there was no estimation involved in annotating the images for this experiment. The annotations were then used to generate the dataset knowledge graph. Each of the datasets consisted of 504 images.

As discussed previously, an ML model's training dataset should be representative of real-world data and real-world proportions of data. The proportions of real-world emergency road vehicles can vary by city, county, or state, so determining the exact proportions of the data used in this experiment was outside the scope of this research. This research assumes that the distribution is equal and that there is a corresponding real life dataset.

*5.1.1. Data Augmentation*

As previously mentioned, some of the images found from Google Images were manually transformed to simulate varying image quality characteristics. This was done using Python code with the libraries listed below to maintain a high level of control over the values of the image characteristics. This also made annotating the images with their characteristics more precise since the values were already known.

More specifically, the following characteristics were programmed using Python's OpenCV library and the image augmentation library and were introduced into the three datasets via this augmentation method:

- Defocus blur using cv2.blur [92]
- Gaussian blur using cv2.GaussianBlur [92]
- Haze blur imgaug.augmenters.weather.Fog [93]
- Motion blur using cv2.filter2D [92] [94]
- Contrast using cv2.convertScaleAbs [95]
- Low daytime illumination using cv2.cvtColor [96]

The resolution of each image was also found using a Python program using the cv2 library [97]. There were a few instances of defocus blur, haze blur, motion blur, low contrast and low daytime illumination within the images chosen from Google images. However, most of the instances used in this experiment that had these characteristics were augmented.

*5.1.2. Dataset One – Full*

The full dataset included enough instances of all the domain entities and image characteristics and had nothing flagged by the process. This dataset was used as the control for this experiment. The intended domain breakdown of the images in the full dataset can be seen in Table I.

As can be seen in Table I, each of the six categories of emergency road vehicles were intended to have an even number of images within the training dataset to avoid representation bias. The image quality characteristics were handled slightly differently because many of the instances were intended to be clear and without low quality characteristics. Additionally, some quality characteristics were expected to occur more often in an autonomous driving application, so those characteristics were prioritized to have more instances. For example, it was expected that motion blur would occur more often than gaussian blur. However, it was important to ensure that each category of quality characteristic was present in each dataset. The actual contents of the dataset will be discussed in more detail later in this paper when discussing the results of the SPARQL queries.

TABLE I
INTENDED DOMAIN BREAKDOWN OF DATASET ONE

| Entity | Number of Instances |
|---|---|
| EMS vehicle | 84 |
| Fire vehicle | 84 |
| Mobile communications vehicle | 84 |
| Police vehicle | 84 |
| Rescue vehicle | 84 |
| Tow truck | 84 |

*5.1.3. Dataset Two – Missing Domain Entity*

The second dataset compiled for this experiment represented a training dataset that was missing a domain entity. In this case, tow trucks were omitted from the dataset, which would be flagged as a source of bias by the SPARQL queries. The intended domain breakdown of the images in this dataset can be seen in Table II.

TABLE II
INTENDED DOMAIN BREAKDOWN OF DATASET TWO

| Entity | Number of Instances |
|---|---|
| EMS vehicle | 101 |
| Fire vehicle | 101 |
| Mobile communications vehicle | 101 |
| Police vehicle | 101 |
| Rescue vehicle | 100 |
| Tow truck | 0 |

As seen in Table II, towing trucks were omitted from this dataset. To maintain the same total number of images within this dataset as in the full dataset, an additional 84 instances of the other categories were included in this dataset. Keeping the same total number of images required having one fewer instance of one type of vehicle in this dataset, in this case, one fewer instance of rescue vehicle than all other types of emergency road vehicles. However, given the scale of the dataset, this one instance was considered negligible. The domain breakdown and the quality characteristic breakdown will be analyzed in more detail with the SPARQL query results.

*5.1.4. Dataset Three – Missing Quality Entity*

The third dataset was compiled to show how the process would flag a dataset that was missing a quality entity. In this experiment, images of haze blur (fog) were omitted from the dataset. The intended domain breakdown for this dataset can be seen in Table III.

Table III shows that each of the categories of emergency road vehicles were intended to be represented evenly in this dataset, just like the first dataset. However, all instances of haze blur were omitted in this dataset.

TABLE III
INTENDED DOMAIN BREAKDOWN OF DATASET THREE

| Entity | Number of Instances |
|---|---|
| EMS vehicle | 84 |
| Fire vehicle | 84 |
| Mobile communications vehicle | 84 |
| Police vehicle | 84 |
| Rescue vehicle | 84 |
| Tow truck | 84 |

*5.2. Dataset Annotation*

Once each of the datasets were compiled, the data needed to be annotated for training and testing models in the experiment. Annotating, or labeling, the data is a method used for supervised learning. By putting labels on the training images, the ML model trained on that data can associate the labels with similar unlabeled instances after training [98]. The annotations were done using object detection in Roboflow [99]. The annotations consisted of labeling not only the types of emergency vehicles within the image, but also each of the quality characteristics of the image. An example annotation can be seen in Fig 18 where there are bounding boxes showing the instances of police vehicles within the image, but also bounding boxes around the entire image to label the quality characteristics. Once annotated, Roboflow can generate datasets already split into training, validation, and testing sets and export the images with their labels. These annotated datasets were then used to train three separate ML models to evaluate the effectiveness of the proposed process. The results of this experiment will be discussed later in this paper.

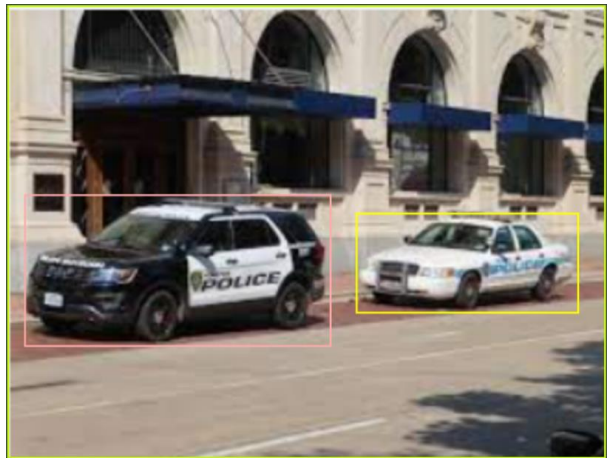

**Fig 18.** Example annotation in Roboflow.

*5.3. Knowledge Graph Creation*

In order to gather domain and quality information from the hundreds of labeled data instances and to compare their labels with the domain and quality ontologies, the labels needed to be formatted into triples. To do this, the images with their labels were exported into a Comma Separated Value (CSV) format, which tied the image file name with all of the labels associated with that image. From there, a tool called Tarql was used to translate the CSV file into triples [100]. Tarql works by feeding the CSV file through a SPARQL query to recognize the subjects and objects and relate them to each other using specified predicates. These triples form the knowledge graphs used for this experiment. They follow the same format as the domain and quality ontologies, with images having subjects and quality characteristics. However, they hold instance data about specific images and are therefore knowledge graphs. Once the CSV files were converted into triples representing the images and all of their labels, the triples were imported into GraphDB, which is a tool for visualizing instance data [101].

The three knowledge graphs made during this part of the experiment were then queried to gather information on the representation and completeness of the datasets. This part of the process is when misrepresentation and bias is flagged.

### 5.4. SPARQL Queries

The purpose of building the knowledge graphs was to put the image labels into a searchable format so that additional knowledge about each of the datasets could then be inferred. Using the SPARQL query function in GraphDB, each of the three knowledge graphs were queried to find the number of images with each category of emergency road vehicle present in them. Additional queries were then run to find the number of images with each type and range of quality characteristic. An example SPARQL query can be seen in Fig 19, which is querying the knowledge graph for the number of images that have a Tow_Truck label. The following subsections will describe the full SPARQL results for each of the datasets. The purpose of these SPARQL queries was to determine if each dataset had enough instances of each class for a user to trust that the model had the opportunity to learn those features.

```
SELECT (COUNT(?s) AS ?triples)
WHERE {
    ?s bfo:BFO_0000051 domain:Tow_Truck .
}
```

**Fig 19.** Example SPARQL query to count the number of images with a Tow_Truck label.

### 5.4.1. Dataset One – Full

After running the SPARQL queries for each category of emergency road vehicles, the results were summarized in Table IV. As can be seen in the table, the breakdown of the domain categories was not perfectly equal as was intended. Although there were still 504 total images in the dataset, some images had more than one emergency road vehicle present in them, and some had more than one type of emergency road vehicle present. While this was initially considered a positive attribute for dataset diversity, the SPARQL queries show that it could affect the representation bias of the dataset. This experimental data speaks to the difficulty in compiling an unbiased dataset for ML and the need for a process to identify such bias.

TABLE IV

EXPERIMENTAL DOMAIN BREAKDOWN OF DATASET ONE

| Entity | Number of Instances |
|---|---|
| EMS vehicle | 103 |
| Fire vehicle | 93 |
| Mobile communications vehicle | 84 |
| Police vehicle | 92 |
| Rescue vehicle | 84 |
| Tow truck | 84 |

TABLE V

EXPERIMENTAL QUALITY CHARACTERISTIC BREAKDOWN OF DATASET ONE

| Entity | No. of Instances (None) | No. of Instances (Low) | Number of Instances (Medium) | Number of Instances (High) |
|---|---|---|---|---|
| Defocus blur | 435 | 63 | N/A | 6 |
| Gaussian blur | 482 | 16 | N/A | 6 |
| Haze blur | 484 | 11 | N/A | 9 |
| Motion blur | 468 | 30 | N/A | 6 |
| Contrast | N/A | 54 | N/A | 450 |
| Illumination | (Night) 54 | (Day, low) 121 | N/A | (Day, high) 329 |
| Occlusion | 307 | 107 | 42 | 48 |
| Resolution | N/A | 1 | 3 | 499 |

SPARQL queries were also performed for the quality characteristics of each image in the dataset. The results of these queries can be found in Table V. As can be seen in the table, these image characteristics are also not completely equal, but the dataset does contain instances from each valid quality category.

*5.4.2. Dataset Two – Missing Domain Entity*

The experimental domain breakdown of Dataset Two can be found in Table VI. As can be seen, the breakdown of the domain categories was again, not equal since many of the images from the first dataset were reused for this dataset. The quality characteristics breakdown can be seen in Table VII. The quality breakdown is similar to that of Dataset One. It is only slightly different as some of the augmented tow truck images were exchanged for some augmented images of other emergency road vehicles.

TABLE VI
EXPERIMENTAL DOMAIN BREAKDOWN OF DATASET
TWO

| Entity | Number of Instances |
|---|---|
| EMS vehicle | 123 |
| Fire vehicle | 111 |
| Mobile communications vehicle | 98 |
| Police vehicle | 106 |
| Rescue vehicle | 98 |
| Tow truck | 0 |

TABLE VII
EXPERIMENTAL QUALITY CHARACTERISTIC
BREAKDOWN OF DATASET TWO

| Entity | Number of Instances (None) | Number of Instances (Low) | Number of Instances (Medium) | Number of Instances (High) |
|---|---|---|---|---|
| Defocus blur | 421 | 65 | N/A | 5 |
| Gaussian blur | 464 | 24 | N/A | 4 |
| Haze blur | 471 | 11 | N/A | 10 |
| Motion blur | 449 | 37 | N/A | 6 |
| Contrast | N/A | 53 | N/A | 439 |
| Illumination | (Night) 56 | (Day, low) 128 | N/A | (Day, high) 308 |
| Occlusion | 302 | 96 | 44 | 50 |
| Resolution | N/A | 0 | 3 | 489 |

*5.4.3. Dataset Three – Missing Quality Entity*

The experimental domain breakdown for this third dataset can be seen in Table VIII. As seen in the table, tow trucks were re-introduced in this dataset. Again, not all of the categories were equal. Table IX shows the experimental quality breakdown for this dataset. Although quite similar to the quality breakdowns for the previous two datasets, it is important to note that for this dataset, all 504 images have no haze blur. This is the representation of a missing quality entity within a dataset for this experiment.

TABLE VIII
EXPERIMENTAL DOMAIN BREAKDOWN OF DATASET
THREE

| Entity | Number of Instances |
|---|---|
| EMS vehicle | 105 |
| Fire vehicle | 92 |
| Mobile communications vehicle | 84 |
| Police vehicle | 91 |
| Rescue vehicle | 83 |
| Tow truck | 84 |

TABLE IX
EXPERIMENTAL QUALITY CHARACTERISTIC BREAKDOWN
OF DATASET THREE

| Entity | Number of Instances (None) | Number of Instances (Low) | Number of Instances (Medium) | Number of Instances (High) |
|---|---|---|---|---|
| Defocus blur | 433 | 65 | N/A | 6 |
| Gaussian blur | 474 | 24 | N/A | 6 |
| Haze blur | 504 | 0 | N/A | 0 |
| Motion blur | 465 | 33 | N/A | 6 |
| Contrast | N/A | 40 | N/A | 464 |
| Illumination | (Night) 54 | (Day, low) 126 | N/A | (Day, high) 324 |
| Occlusion | 308 | 107 | 39 | 50 |
| Resolution | N/A | 1 | 3 | 499 |

*5.5. Experiment Result*

With the three datasets compiled and annotated, the ML models could then be trained and tested. Three models were trained via transfer learning, one for each of the datasets, so that the models' performance could be compared to measure the effectiveness of the process. The following subsections will go into detail on the results of the model testing and the next section provides analysis on the results. The models trained for this experiment were YOLOv8 models [102] [103]. Each of the models was trained on a 504-image dataset for 100 epochs with patience set to 50. It is expected that larger datasets would improve the overall performance of the models. However, the main goal of the experiment was to analyze the effect of the identified missing entities within Datasets Two and Three, and therefore, the effectiveness of the process. With this main goal in mind, the overall performance of the models was not prioritized.

*5.5.1. Model One – Full*

After training on the full dataset for 100 epochs, the first YOLOv8 model had a precision of 0.41 and a recall of 0.59. Additional data would be expected to increase model performance. However, the main purpose of this experiment was to analyze the difference in performance specifically for the missing entities in the other datasets, so this model's performance was a good baseline for this experiment.

To analyze this performance difference, a special test set was compiled with images of only the missing entities. The exact predictions and confidence levels of the first model on this test set can be seen in the Appendix where empty prediction and confidence cells indicate that the model did not assign any class to the object in the image. In total, 15 out of the 16 tow trucks in the images were correctly identified and many were identified with over 0.80 confidence level. This performance will be compared with the second model, which was trained on the dataset missing images of tow trucks. Images 16-30 in this test set all include low or high haze blur, or fog level. The performance of the m odel on these images is much worse than on the first 15 images, with only 10 out of 15 correctly labeled and some with much lower confidence levels. However, there were far fewer instances of haze blur in the dataset than there were tow trucks, suggesting that increasing the number of instances of haze blur in the training dataset could solve this problem.

*5.5.2. Model Two – Missing Domain Entity*

The second model was trained on Dataset Two, which was missing a domain entity. The overall precision and recall were 0.38 and 0.46, respectively. The predictions and confidence levels of the second model on the special test set can be seen in the Appendix. In contrast to Model One, this model did not correctly identify any of the tow trucks in this test set. This vast difference in performance indicates that the absence of tow trucks in the training data reduces the model's performance on tow trucks. This experimental result suggests that the process that flagged the absence of tow trucks in the second dataset would work in improving the safety of the model's training dataset if the developer then added tow trucks into the dataset.

*5.5.3. Model Three – Missing Quality Entity*

The final model was trained on Dataset Three, which omitted all instances of haze, or fog. The overall precision and recall for this model were 0.74 and 0.46, respectively. The predictions of Model Two on the special test set can be seen in the Appendix. This model correctly predicted 6 out of 15 of the hazy images, which indicates that the system was again successful in identifying an absence in the training dataset. Although there was not quite the same amount of success as in the domain experiment, this could be attributed to not having as many hazy instances in the full training set as is needed for good performance. It is also important to note that the third model performed much better on test instances with low haze blur, rather than high haze blur whereas the full model was able to handle both.

*5.6. Analysis of Results*

As seen in the previous section, Model One, trained on the full dataset, showed a remarkable improvement in its identification of tow trucks over Model Two, which was trained on the dataset missing all instances of tow trucks. These results suggest that the process, which flagged Dataset Two for not including any instances of tow trucks based on the desired uniform class distribution, would improve the robustness of an ML model's training dataset if the developer using the process, then changed the dataset accordingly.

Model Three, trained on the dataset missing instances of haze blur, did not show as much difference in accuracy than Model One. This is likely due to not having enough instances of haze blur within the full dataset for the first model to show improvement over Model Three. This suggests that an improvement to the process is needed to ensure that enough instances of each of the quality characteristics are present in a dataset. However, both experiments were successful, showing the usefulness of this proposed system.

## 6. Related Work

Several papers discuss the importance of the training dataset in making a safe ML model [6]. This can be caused by representation bias, which is overrepresenting certain classes or characteristics and underrepresenting others [6]. To fix this issue, some suggest making synthetic data instances to increase the number of minority instances [104] [105]. This method is often cheaper than collecting real-world instances and can fill diversity gaps in the original dataset [104]. However, current techniques that apply this mitigation method use decision trees, naïve Bayes classifiers, neural networks, or support vector machines as ways to classify the dataset gaps [105]. In contrast, this research suggests using the semantic web to identify gaps in a dataset by comparing the characteristics found in the dataset to all characteristics found in a domain ontology.

Another way to approach the issue of representation bias is to use semi-supervised ML models, which utilize a small proportion of labeled data and a much larger proportion of unlabeled data for training [106]. Semi-supervised learning relies on clustering and predicting based on proximity to a cluster [106]. This allows for datasets to be larger because of the amount of available unlabeled data there is [106]. However, this technique assumes that the dataset is sufficiently varied because of its size, rather than checking the characteristics of each data instance to find underrepresented characteristics. This research compares the data to domain and quality characteristic ontologies to ensure sufficient representation of all entities.

Another paper suggests that one way to ensure safety in a model is to understand which data partitions have sufficient data to make a safe decision and which do not [107]. By doing this, the model can then only be allowed to decide in the partitions that have enough training data [107]. Although this method does not address the bias, it takes a good first step in doing so by identifying what parts of the dataset have insufficient data, which would indicate bias. However, this reduces the utility of the model and applies safety measures after training by suppressing any decision without high probability of classification within a safe partition. In contrast, this research applies safety measures prior to training to ensure the safety of the model, rather than safety of each individual decision.

More recently, there has been a publication on using a domain ontology to check the completeness of an ML training dataset [108]. This paper built a domain ontology for the hard-to-specify domain of pedestrians using Google books N-gram, RelatedWords, Google News, and other knowledge bases. After building the domain ontology, they trained two ML models, one with pedestrians in wheelchairs, and one without to show that using the domain ontology to build the training dataset would help make the model more accurate in identifying pedestrians. The work of these authors is similar to the domain ontology piece of this research. However, the research presented here includes an image quality characteristic ontology to improve ML model robustness to varying quality characteristics. This research also includes an ML ontology, and all of the ontologies are aligned with BFO to ensure their standardization and easy expansion and reuse. The research presented here also stores all of the images in the training dataset in a knowledge graph, which allows the instance data to be queried for easy access of information about it, which the referenced paper does not do.

Another recent publication focuses on the collection of domain knowledge to refine requirements engineering around ML training datasets [109]. The paper argues that there needs to be a standard process to define the domain of a training dataset so that the requirements can be written, especially for safety-critical ML applications. While the motivation of the paper matches the motivation of this research, the methods vary and this research includes an image quality characteristic check, which the referenced paper does not.

## 7. Threats to Validity

A threat to the validity of the research results presented in this paper is that the datasets used in training the ML models did not have balanced classes of domain entities as was intended. This speaks to the difficulty of balancing training datasets and the need for a validation process because even with effort put into balancing the data, the datasets fell short of that.

Additionally, it was assumed that a balanced dataset with respect to the domain classes had a corresponding dataset in the real world. While this problem was outside of the domain, the assumption should be addressed further in future work via domain research to ensure the proper distribution of emergency vehicles.

Finally, the overall performance of the models trained on these datasets was relatively poor compared to the performance of many ML models today. Although this was attributed to the size of the training datasets, this problem should be addressed in future work.

## 8. Future Work

The results presented in the previous sections are promising. However, as stated in the results section, they should be confirmed. Given that the datasets were slightly imbalanced with respect to the domain classes, the experiment should be repeated with fully balanced domain classes to match the goal of the experiment. Additionally, it could be beneficial to repeat the experiment with more images in each dataset and with more epochs during model training to improve the overall performance of the models.

The domain ontology could also be aligned with an upper-level ontology, called the Common Core Ontologies, which describes many basic real-world artifacts, such as vehicles [110]. There are also many more image quality characteristics that would be useful in a training dataset for use in autonomous vehicles, such as images with reflections, rain, or snow.

It would also be beneficial to automate this process. The process can be tedious, time-consuming, and expensive. As seen from the results, even after effort and time spent on balancing the datasets, mistakes can still be made. Automating the process of building the domain ontology via the semantic web, running the training data through the SPARQL queries, and providing insights into the biases would make the process faster and less error prone. Automatically fixing the dataset would also be beneficial in the direction of future work. This could be done by making synthetic instances of data to fill in gaps in the dataset and simply return a balanced and complete dataset to the developer.

This paper focused on object detection models, but the experiment could also be run on other types of ML models, such as Natural Language Processing (NLP). However, the method presented here must still be used for supervised learning because the knowledge graphs are built using the labels.

One of the goals of this research was to move towards human-out-of-the-loop ML in safety-critical domains. The application was for autonomous driving due to ease of collecting data. However, autonomous driving is already human-out-of-the-loop, so future research could explore other safety-critical domains that are still human-in-the-loop, such as cancer diagnosis.

## 9. Conclusions

This work proposes a solution to the lack of user trust in ML models in safety-critical applications. In applications where life and livelihood are at risk, it is essential that ML models do not exhibit biased traits. Existing performance metrics and explainability methods do not provide enough confidence that the model is making fair and balanced decisions. This confidence is difficult to build after an ML model has already been trained because at that point, explainability methods are only ensuring the safety of specific decisions, which limits the ability to move to human-out-of-the-loop ML since a domain expert still has to review every decision. However, if a training dataset can be shown to be safe, this can increase the trust of users in a model trained on that data.

The work presented in this paper describes a method of ensuring the robustness and completeness of an ML training dataset. It includes a domain ontology to measure the domain completeness of the training dataset, as well as an image quality characteristic ontology to measure the expected robustness that the model will have from that training dataset. Furthermore, it includes an experiment within the domain of recognizing emergency road vehicles during autonomous driving. The first part of the experiment compared a model trained on a full dataset with one trained on a dataset missing all instances of towing trucks. The second part of the experiment compared the full model with one missing all instances of haze blur, or fog. Potential areas of future work are to confirm the results from this research, align the domain ontology to an upper-level ontology, and automate the process to make the wide-spread use of such a process easier and less time-consuming for dataset engineers.

## Appendix

Table X shows the individuals and their values for the image quality characteristic ontology.

To view each of the four datasets used to train the ML models in the experiments, as well as expansions of the ontologies and results of the experiments, please visit https://github.com/lynndalou/SafeDatasets/tree/main.

TABLE X

SMALL CAPS: INDIVIDUALS AND VALUE RANGES FOR THE IMAGE QUALITY CHARACTERISTIC ONTOLOGY

| Individual | Definition (Value Range) | Reference |
|---|---|---|
| **Occlusion** | | |
| Occlusion_None | 0% occluded | |
| Occlusion_Low | 1 − 40% occluded | *Compositional Convolutional Neural Networks: A Deep Architecture with Innate Robustness to Partial Occlusion* [112] |
| Occlusion_Medium | 40 − 60% occluded | *Compositional Convolutional Neural Networks: A Deep Architecture with Innate Robustness to Partial Occlusion* [112] |
| Occlusion_High | 60 − 80% occluded | *Compositional Convolutional Neural Networks: A Deep Architecture with Innate Robustness to Partial Occlusion* [112] |
| **Contrast** | | |
| Contrast_Low | 0 − 0.5 relative luminance | *Benchmarking Neural Network Robustness to Common Corruptions and Perturbations* [113]<br><br>*Local Luminance and Contrast in Natural Images* [114] |
| Contrast_High | 0.5 − 1 relative luminance | *Benchmarking Neural Network Robustness to Common Corruptions and Perturbations* [113]<br><br>*Local Luminance and Contrast in Natural Images* [114] |
| **Defocus_Blur** | | |
| Defocus_Blur_None | 0 pixels | |
| Defocus_Blur_Low | 1 − 15 pixels | *Recognition of Images Degraded by Linear Motion Blur without Restoration* [115] |
| Defocus_Blur_High | 15 − 30 pixels | *Recognition of Images Degraded by Linear Motion Blur without Restoration* [115] |
| **Gaussian_Blur** | | |
| Gaussian_Blur_None | 0 pixels | |
| Gaussian_Blur_Low | 1 − 15 pixels | *Recognition of Images Degraded by Linear Motion Blur without Restoration* [115] |
| Gaussian_Blur_High | 15 − 30 pixels | *Recognition of Images Degraded by Linear Motion Blur without Restoration* [115] |
| **Haze_Blur** | | |
| Haze_Blur_None | 0 pixels | |
| Haze_Blur_Low | 1 − 15 pixels | *Recognition of Images Degraded by Linear Motion Blur without Restoration* [115] |
| Haze_Blur_High | 15 − 30 pixels | *Recognition of Images Degraded by Linear Motion Blur without Restoration* [115] |
| **Motion_Blur** | | |
| Motion_Blur_None | 0 pixels | |
| Motion_Blur_Low | 1 − 15 pixels | *Recognition of Images Degraded by Linear Motion Blur without Restoration* [115] |
| Motion_Blur_High | 15 − 30 pixels | *Recognition of Images Degraded by Linear Motion Blur without Restoration* [115] |
| **Illumination** | | |
| Illumination_Night | < 1,000 lux | *Indoor Versus Outdoor Time in Preschoolers at Child Care* [116]<br><br>*The Effects of Different Outdoor Environments, Sunglasses, and Hats on Light Levels: Implications for Myopia Prevention* [117] |
| Illumination_Day_Low | 1,000 − 11,000 lux | *Indoor Versus Outdoor Time in Preschoolers at Child Care* [116]<br><br>*The Effects of Different Outdoor Environments, Sunglasses, and Hats on Light Levels: Implications for Myopia Prevention* [117] |
| Illumination_Day_High | > 11,000 lux | *Indoor Versus Outdoor Time in Preschoolers at Child Care* [116]<br><br>*The Effects of Different Outdoor Environments, Sunglasses, and Hats on Light Levels: Implications for Myopia Prevention* [117] |
| **Resolution** | | |
| Resolution_Low | 32x32 − 64x64 | *Impact of Image Resolution on Deep Learning Performance in Endoscopy Image Classification: An Experimental Study Using a Large Dataset of Endoscopic Images* [118] |
| Resolution_Medium | 64x64 − 256x256 | *Impact of Image Resolution on Deep Learning Performance in Endoscopy Image Classification: An Experimental Study Using a Large Dataset of Endoscopic Images* [118] |
| Resolution_High | 256x256 − 512x512 | *Impact of Image Resolution on Deep Learning Performance in Endoscopy Image Classification: An Experimental Study Using a Large Dataset of Endoscopic Images* [118] |